# Designing a Wayfinding Robot for People with Visual Impairments

Shuijing Liu*, Aamir Hasan*, Kaiwen Hong, Chun-Kai Yao, Justin Lin, Weihang Liang,
Megan A. Bayles, Wendy A. Rogers, and Katherine Driggs-Campbell

*Abstract*— People with visual impairments (PwVI) often have difficulties navigating through unfamiliar indoor environments. However, current wayfinding tools are fairly limited. In this short paper, we present our in-progress work on a wayfinding robot for PwVI. The robot takes an audio command from the user that specifies the intended destination. Then, the robot autonomously plans a path to navigate to the goal. We use sensors to estimate the real-time position of the user, which is fed to the planner to improve the safety and comfort of the user. In addition, the robot describes the surroundings to the user periodically to prevent disorientation and potential accidents. We demonstrate the feasibility of our design in a public indoor environment. Finally, we analyze the limitations of our current design, as well as our insights and future work. A demonstration video can be found at https://youtu.be/BS9r5bkIass.

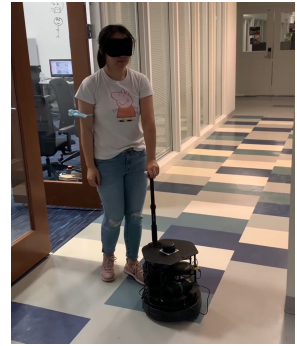

Fig. 1: Overview of the robot platform and the navigation configuration.

## I. INTRODUCTION

Finding a restroom in a building given simple directions and signages is easy for healthy individuals yet especially difficult for people with visual impairments (PwVI). PwVIs usually have to rely on a companion or white canes to navigate [1], [2]. As a result, they tend to avoid going to unfamiliar places, which drastically reduces the quality of their life. Motivated by this need, researchers have used robots to plan paths and guide PwVIs to their desired goals [3]–[5]. However, some of these methods make assumptions on the position of the user with respect to the robot, and fail to account for the inevitable changes in the position of the user during navigation. In addition, these pipelines ignore communications with the user, such as providing important information about the environment during navigation.

To address the above problems, we propose to incorporate user pose estimation into path planning for an assistive indoor wayfinding system. After receiving a verbal command from the user, the robot plans a path and guides the user to navigate to the destination, as shown in Fig. 1. By using the robot arm as the main point of contact, our design provides both directions and extra support to the user. Meanwhile, the robot adjusts its plans based on the estimated pose of the user to avoid potential collisions of the user. To prevent spatial disorientation of the user, the robot also describes the surrounding environments periodically with a speaker. Our goal is to allow the user to navigate safely and efficiently without a companion even in unfamiliar indoor spaces.

## II. METHODOLOGY

In this section, we describe the setup and configuration of our robot guide, as shown in Fig. 1. We also describe the modules contained in our design: the estimation of the user's pose, the planning module with the estimated pose, scene description, and audio communication.

### A. Robot platform

We use the Turtlebot2i robot as our robot platform. The Turtlebot2i is a lightweight robot with a mobile base that can be modified to include hardware devices and sensors as needed. Our robot is fitted with the following sensors: (1) An RP-Lidar A3 laser range finder is mounted at the top of the robot structure to perform simultaneous localization and mapping (SLAM) for navigation and user detection and pose estimation; (2) Intel RealSense D435i camera is mounted atop a extender on the robot to detect the user and estimate their position and orientation; (3) A microphone array is used for the robot to communicate with the user and placed below the Lidar mount; A selfie stick was mounted on top of the robot structure to serve as a holding point for the user's arm.

### B. Audio communication

A user study finds that some PwVI prefer to have additional knowledge of their environment during navigation to maintain a mental model of their surroundings and locate themselves [2]. Additionally, the participants in the same study mentioned feeling most comfortable while verbally communicating with a robot over methods such as using phone apps or buttons. Hence, we developed an audio communication module that takes in the users input to navigate them to a requested destination, in addition to communicating the output from the scene description module. Before the navigation begins, our current module uses a simple speech-to-text extractor [6] with a bag-of-words model to extract the requested destination within the testing location. We use a text-to-speech library to utter the output sentences from the scene description module in Sec. II-E using a speaker on the robot [7]. To further improve user experience our module also reiterates the users requested destination before beginning navigation.

*C. User pose estimation*

The usual robot navigation pipeline only considers the collision avoidance of the robot but not the user, which will result in collisions or discomfort of the user. For this reason, we extract the user's position and orientation in the 2D Euclidean space with body segmentation from depth images. First, we use BodyPixs, an open-source machine learning model, with MobileNet [8] to segment the user's torso on the RGB image from the background. Then, we use the depth values of the torso segment on the depth image to estimate the user's y-coordinate and facing angle in the camera frame using trigonometry. For the x-coordinate, we first calibrate the camera parameters, which are used to approximate the offset distance of the human torso from the center of the camera and compute the x-coordinate of the user. In the last step, we treat the boundary of the user as a rectangle and compute the coordinates of the vertices of the rectangle in the camera frame with a homogeneous transformation between the camera and the user frame.

*D. Pose-aided Navigation*

Before navigation, the robot maps the environment using simultaneous localization and mapping (SLAM) with Gmapping ROS package [9] During SLAM, the robot estimates its location using its wheel odometry, LiDAR scans, and adaptative Monte Carlo localization (AMCL) [10]. Given the extracted destination from the audio module, our program finds the corresponding goal pose and passes it to the planner. To plan a path for the robot-human team based on the map, we use the A* algorithm to plan a global path and the dynamic window approach (DWA) as the local path planner [11]. To incorporate pose information into the DWA planner, we change the collision boundary of the robot so that it also includes the estimated boundary of the user from Sec. II-C. More specifically, given the estimated pose of human, we generate a boundary box on the user's reachable spaces and use a polygon to represent the box in the robot frame, which are fed into the DWA planner. As a result, the planned paths avoids collisions for both the user and the robot in real-time.

*E. Scene description*

Based on a recent user study, PwVIs frequently get disoriented during navigation, especially in large and open spaces [1]. To help PwVIs remember the layout of the space and locate themselves, our design uses a state-of-the-art image captioning method [12] to inform the user of its surroundings. The image captioning model takes an RGB image from the camera as the input, and outputs a sentence that describes the main objects and their relations in the image. The model is a deep neural network trained using a dataset of images and their captions. This process repeats periodically until the end of the navigation.

## III. PRELIMINARY EXPERIMENTS

Our experiments are conducted in a university building that mimics an unfamiliar indoor environment. We test the feasibility of our design with a blindfold person. In our experiments, to receive accurate real-time guidance, the user held on to the selfie stick with a perpendicular configuration as shown in Fig. 1. A demonstration video can be found at https://youtu.be/BS9r5bkIass. In the video, after receiving a verbal command from the user, the robot successfully plans a path to the user-specified goals and executes the path smoothly without collisions or severe jerks.

## IV. LIMITATIONS AND FUTURE WORK

Though our design addresses gaps in previous attempts at designing wayfinding robots, it posses the following limitations. (1) Due to the close proximity of the user to the sensors on the robot, the pose estimation module might be inaccurate which could potentially lead to collisions between the user and the mobile base of the robot. Adding additional sensors near the base of the robot could help mitigate this issue. (2) Since PwVIs use auditory cues to perceive their environment, the communications of our robot with the user may interfere this process. (3) Natural language communication can be too slow for events that require a quick reaction. A possible solution is to include tactile sensors or use tones to communicate potential hazardous conditions.

For future work, we plan on improve our model to enhance the users experience with the robot in the following ways. First, during navigation, we will use data-driven methods to predict user's motion when the robot is performing a turn or a sudden stop to improve the comfort and safety of the planned paths. Second, for the scene description module, we plan to give higher priorities to objects that are potentially more dangerous to the user such as moving people and pets. In addition, we plan to use other sensors such as depth images as complements to enrich the information in the descriptions. Third, for the audio communication module, we plan to replace the bag-of-words model with a learning-based model to improve the extraction of the user's intent from more complex sentences.

## V. CONCLUSIONS

In this short paper, we present our current design for a wayfinding robot for PwVI with different modules and propose improvements to make the navigation process more comfortable for the user. Our experiments show the feasibility of such a robot and provide a strong base for a promising solution to aide PwVI in navigating unfamiliar environments. After the proposed improvements are added, we plan on conducting a user study where PwVI use our robot to explore an unfamiliar space.